%% file: main.tex
\documentclass[10pt,twocolumn,letterpaper]{article}
\usepackage[whole]{bxcjkjatype} 

\usepackage{cvpr}              
\input{preamble}
\definecolor{cvprblue}{rgb}{0.21,0.49,0.74}
\usepackage[pagebackref,breaklinks,colorlinks,allcolors=cvprblue]{hyperref}
\usepackage[ruled,vlined]{algorithm2e}
\usepackage{multirow}
\usepackage{multicol}
\usepackage{array}
\usepackage{booktabs}
\usepackage{arydshln}
\usepackage{float}

\def\paperID{39587} 
\def\confName{CVPR}
\def\confYear{2026}

\title{Teacher-Guided Routing for Sparse Vision Mixture-of-Experts}

\author{
\begin{tabular}{ccccc}
Masahiro Kada\textsuperscript{1} &
Ryota Yoshihashi\textsuperscript{1} &
Satoshi Ikehata\textsuperscript{2,3} &
Rei Kawakami\textsuperscript{1} &
Ikuro Sato\textsuperscript{1,2}\\
\end{tabular}\\
\begin{tabular}{ccc}
\textsuperscript{1}Institute of Science Tokyo &
\textsuperscript{2}DENSO IT Laboratory &
\textsuperscript{3}National Institute of Informatics \\
\end{tabular}\\
Tokyo, Japan\\
{\tt\small kada@d-itlab.comp.isct.ac.jp}
}

\usepackage{xcolor}
\newcommand{\fix}[1]{#1}

\begin{document}
\maketitle
\input{sec/0_abstract}
\input{sec/1_intro}

\input{sec/2_relatedwork}
\input{sec/3_methods}

\input{sec/4_evaluation}
\input{sec/5_analysis}
\input{sec/6_conclusion}

\section{Acknowledgement}
This work was supported by DENSO IT LAB Recognition, Control, and Learning Algorithm Collaborative Research Chair. All experiments were carried out using the TSUBAME4.0 supercomputer at Institute of Science Tokyo.
{
    \small
    \bibliographystyle{ieeenat_fullname}
    \bibliography{main}
}

\clearpage
\input{sec/X_suppl}

\end{document}

%% file: sec/0_abstract.tex
\begin{abstract}

Recent progress in deep learning has been driven by increasingly large-scale models, but the resulting computational cost has become a critical bottleneck.
Sparse Mixture of Experts (MoE) offers an effective solution by activating only a small subset of experts for each input, achieving high scalability without sacrificing inference speed.
Although effective, sparse MoE training exhibits characteristic optimization difficulties.
Because the router receives informative gradients only through the experts selected in the forward pass, it suffers from gradient blocking and obtains little information from unselected routes.
This limited, highly localized feedback makes it difficult for the router to learn appropriate expert-selection scores and often leads to unstable routing dynamics, such as fluctuating expert assignments during training.
To address this issue, we propose TGR-MoE: Teacher-Guided Routing for Sparse Vision Mixture-of-Experts, a simple yet effective method that stabilizes router learning using supervision derived from a pretrained dense teacher model.
TGR-MoE constructs a teacher router from the teacher's intermediate representations and uses its routing outputs as pseudo-supervision for the student router, suppressing frequent routing fluctuations during training and enabling knowledge-guided expert selection from the early stages of training.
Extensive experiments on ImageNet-1K and CIFAR-100 demonstrate that TGR consistently improves both accuracy and routing consistency, while maintaining stable training even under highly sparse configurations.
\end{abstract}

%% file: sec/1_intro.tex
\section{Introduction}
\label{sec:intro}

Sparse Mixture-of-Experts (MoE) architectures~\cite{shazeer2017outrageously} enable compute-efficient scaling by activating only a small subset of experts per input, allowing model capacity to grow without a proportional increase in computation. Originally popularized in large-scale language models~\cite{fedus2022switch,lepikhin2020gshard}, MoE has recently been extended to vision models for recognition and generation tasks~\cite{riquelme2021scaling,puigcerver2023sparse,videau2025mixture,Feng_2023_CVPR}. In this work, we focus on \emph{sparse Vision MoE (VMoE)} (\eg,~\cite{riquelme2021scaling,puigcerver2023sparse}), where individual patch tokens are fed into MoE layers in parallel.

\begin{figure}[!t]
  \centering
  \includegraphics[width=1.00\linewidth]{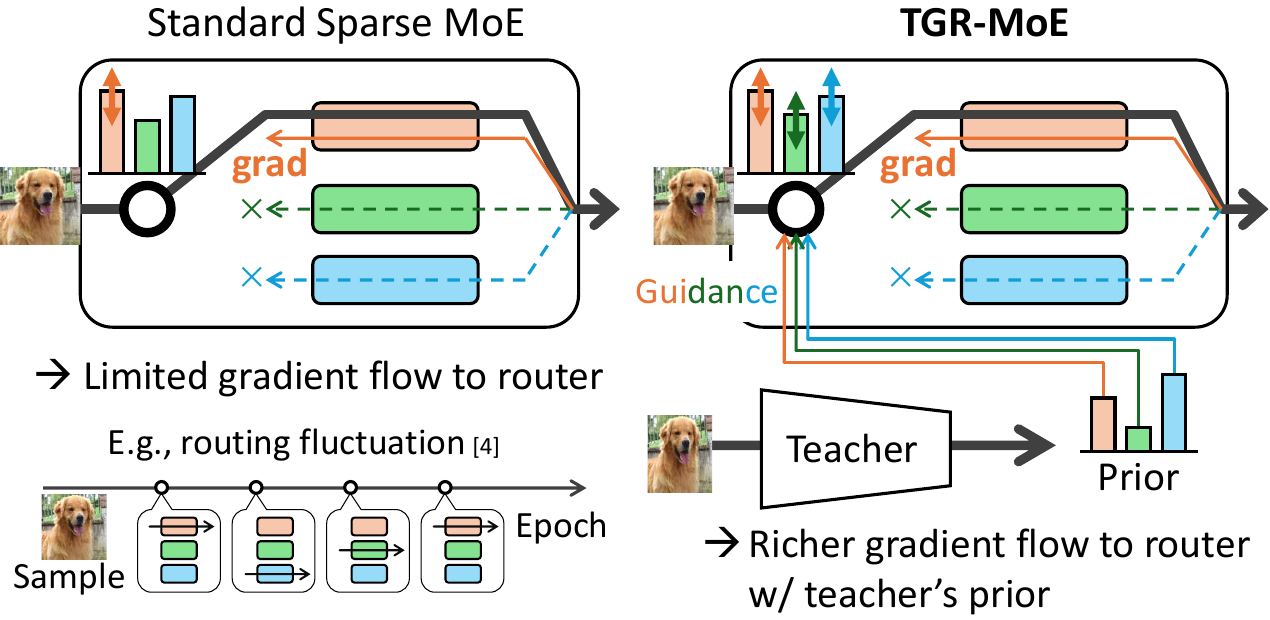}
  \caption{
  Comparison between standard sparse MoE routing (left) and our Teacher-Guided Routing in TGR-MoE (right).
  The bar plots denote the router's per-expert routing scores.
In a conventional sparse MoE, the router receives gradient feedback only from the experts it selects, leaving the majority of experts without informative signals. This limited and highly localized feedback can lead to locally optimal routing behaviors, such as expert collapse or frequent changes in expert selection.
TGR-MoE introduces a global prior derived from the intermediate features of a pretrained dense teacher model. This prior provides additional supervision to the router, complementing the sparse gradient signals and encouraging more coherent routing preferences even for experts that are not selected at each step.
  }
  \label{fig:intro}
\end{figure}

\fix{Despite their computational advantages, sparse MoE models remain challenging to train stably~\cite{zoph2022stmoe}. Unlike dense (\ie, non-MoE) models, which update all parameters on every input, sparse MoE architectures activate and update only a small subset of experts and the router for each input. We argue that an important yet under-discussed challenge in this training regime is \emph{gradient blocking}: as illustrated in Fig.~\ref{fig:intro} (left), because only selected experts participate in the forward and backward passes, the router receives no informative feedback from unselected routes. As a result, it is difficult for the router to learn appropriate expert-selection scores for each input, especially in the early stage of training when expert specialization has not yet emerged.}

\fix{This lack of informative, data-dependent supervision can easily lead to unstable routing dynamics. In particular, the expert assigned to a fixed input can vary in an excessively frequent fashion over the course of training, a phenomenon referred to as \emph{routing fluctuation}~\cite{dai-etal-2022-stablemoe}. Although auxiliary load-balancing losses~\cite{shazeer2017outrageously} are commonly used to mitigate expert-usage imbalance, they are not sufficient to effectively suppress routing instability during training and can even amplify changes in expert assignment over the course of training. Such temporal inconsistency complicates optimization because the same sample may update different experts across iterations, even though only one or a small fixed set of experts will be used at inference. Overly frequent changes in expert assignment hinder stable expert specialization and can leave some experts undertrained, particularly in settings with a large number of experts. These considerations motivate us to provide the router with an external prior that supplies informative guidance beyond the sparsely activated routes, as illustrated in Fig.~\ref{fig:intro} (right).}

We propose \textit{TGR-MoE: Teacher-Guided Routing for Sparse Vision Mixture-of-Experts}, a framework that stabilizes sparse MoE routing by leveraging a pretrained dense teacher. Rather than modifying the task objective, TGR-MoE attaches an auxiliary \emph{teacher router} to the frozen teacher backbone. 
The teacher router is trained with a load-balancing loss and an entropy loss, which were formerly used as regularizers for the target router, 
to produce routing distributions that are both balanced across experts and sufficiently confident. 
The resulting teacher router yields stable routing distributions that are used to guide the target router.
The MoE student learns its router by distilling these distributions: for each MoE layer, the student gating distribution is trained to match the teacher's via a Kullback--Leibler divergence term combined with the standard task loss. Because the teacher backbone is frozen and only the lightweight teacher router is optimized, teacher routing converges quickly and remains stable, and the distillation term encourages temporally consistent, semantically meaningful expert assignments in the student. 
The proposed training strategy introduces a prior to the router's softmax outputs with different abstraction levels, relieving the gradient blocking problem \fix{without freezing the target router}, while no additional cost is required at test time.

We evaluate TGR-MoE against the standard VMoE~\cite{riquelme2021scaling} and its variants. Across ImageNet-1K~\cite{imagenet} pre-training and downstream tasks including CIFAR-10/100~\cite{cifar} and Oxford-IIIT Pets~\cite{pets}, TGR-MoE consistently outperforms VMoE at all model scales (Tiny, Small, Base) under a comparable compute budget. It is competitive with, and often superior to, advanced MoE variants such as expert-choice routing~\cite{NEURIPS2022_2f00ecd7}, SoftMoE~\cite{puigcerver2023sparse}, and z-loss regularization~\cite{zoph2022stmoe}. Applying TGR-MoE during fine-tuning further yields the strongest transfer performance on CIFAR-100, indicating that explicit routing guidance remains beneficial beyond pre-training. TGR-MoE also scales more reliably with the number of experts, maintaining steady gains even in regimes where conventional VMoE saturates (Sec.~\ref{sec:num_experts}), and routing-consistency analysis shows that it substantially reduces routing fluctuations and reaches stable expert assignments earlier than VMoE (Sec.~\ref{sec:analysis_consistency}).

%% file: sec/2_relatedwork.tex
\section{Related Work}
\label{sec:related}

\paragraph{Mixture of Experts.}
While MoE was originally introduced in the early 1990s~\cite{jacobs1991adaptive, NIPS1991_05f971b5},  
it has recently emerged as an efficient way to scale model capacity without increasing computation proportionally~\cite{shazeer2017outrageously, lepikhin2020gshard}.  
Driven by scaling laws~\cite{kaplan2020scaling,chinchilla}, MoE has been widely adopted in large language models~\cite{lepikhin2020gshard,du2022glam,fedus2022switch,lewis2021base,dai2024deepseekmoe,jiang2024mixtral,openmoe} 
and vision models~\cite{riquelme2021scaling,mi2023switchnerf,Feng_2023_CVPR,Cong_2023_ICCV,puigcerver2022adversarial,Zhang_2023_ICCV,NEURIPS2023DAMEX,variational_moe,kada}.
Despite this progress, MoE architectures face several training challenges. A well-known issue is expert-usage imbalance, where only one or a few experts receive most routing traffic; auxiliary load-balancing losses~\cite{shazeer2017outrageously,fedus2022switch,riquelme2021scaling} and alternative designs such as expert-choice routing~\cite{NEURIPS2022_2f00ecd7} or normalized routing strategies~\cite{deepseekai2025deepseekv3technicalreport} aim to address this.
Prior work has also noted numerical sensitivities arising from interactions between router logits and expert outputs~\cite{zoph2022stmoe}. To mitigate issues caused by gradients flowing only through chosen experts, giving the router localized and discontinuous feedback, several methods introduce continuous relaxations of routing, including DSelect-k~\cite{dselectk}, differentiable top-$k$ gating~\cite{sander2023fast}, and ReMoE~\cite{remoe}.
Although these smooth the gating function, expert activations remain discrete.
Fully continuous formulations such as Soft-MoE~\cite{puigcerver2023sparse} eliminate discrete selection entirely, while approaches like SMEAR~\cite{softmerge} merge experts via adaptive soft combinations.

Sparse MoE training also exhibits \emph{routing fluctuation}, where expert assignments vary across training steps.
StableMoE~\cite{dai-etal-2022-stablemoe} and HashMoE~\cite{roller2021hash} improve stability through decoupled or fixed routing, highlighting the value of consistent assignments. However, such methods constrain or freeze routing, limiting the model's ability to adapt routing decisions as representations evolve. In this paper, instead of fixing routing, we guide the router during training with an external, stable source of global guidance.

\vspace{1mm}
\noindent
\textbf{Knowledge Distillation.}
Knowledge distillation (KD) transfers predictive behavior or intermediate representations from a teacher model to a student~\cite{hinton2015distillingknowledgeneuralnetwork,Romero2015FitNets}.  
Representative vision distillation methods include DeiT \cite{pmlr-v139-touvron21a}, which introduced token-based teacher guidance for ViTs, DKD~\cite{zhao2022dkd}, which separates target and non-target logits to stabilize training, and CAT-KD~\cite{guo2023class}, which leverages class-wise attention to enhance feature alignment.
A few recent studies distill information directly into the router, but with objectives fundamentally different from ours.  
Read-ME~\cite{readme} transfers the activation sparsity pattern of a pretrained dense model to an MoE router in order to reproduce the teacher's computation during model conversion, and  
Dynamic Expert Specialization~\cite{desmoe} constrains a fine-tuned router to remain close to the routing distribution of a pretrained MoE to mitigate catastrophic forgetting.  
Both methods therefore apply distillation to preserve a pre-existing routing behavior for purposes such as model conversion or domain adaptation. However, neither approach directly addresses the instability that originates from discrete routing during pretraining.


%% file: sec/3_methods.tex
\section{Method}
\label{sec:methods}
\begin{figure}[t]
    \centering
    \includegraphics[width=1.00\linewidth]{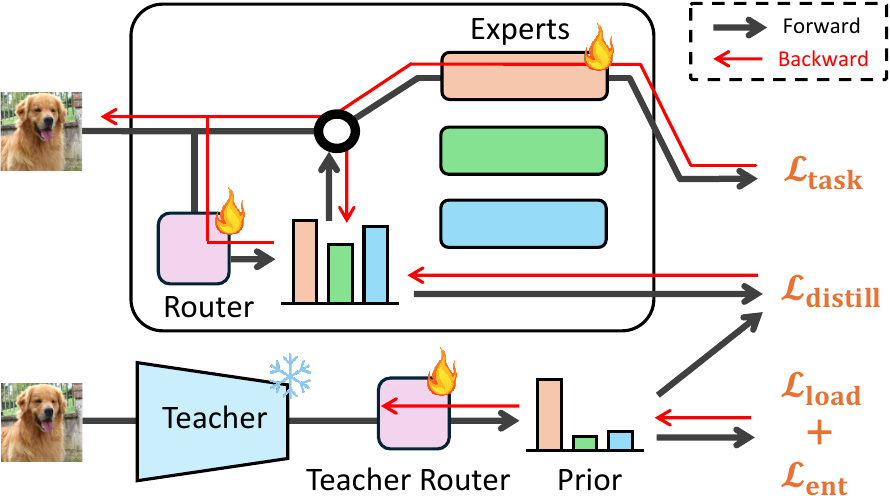}
    \caption{Detailed architecture of the proposed TGR-MoE, illustrated for layer $i$. The teacher provides a routing signal $\mathcal{L}_\text{distill}$ to guide the student router, with the additional balancing loss $\mathcal{L}_\text{load}$ and the regularization loss $\mathcal{L}_\text{ent}$.
}
    \label{fig:overview}
\end{figure}

We propose TGR-MoE, a framework that stabilizes sparse MoE routing by leveraging a pretrained dense teacher model. The teacher router provides a stable routing distribution that supervises the student router during training.

\subsection{Preliminary: MoE Routing}
In Transformer architectures, the MoE layer is typically introduced
as a replacement for the feed-forward network (FFN) within each block.
Each MoE layer consists of $E$ experts $\{f_e(\cdot)\}_{e=1}^{E}$,
each implemented as an MLP with independent parameters.
Given $\mathbf{h}\in\mathbb{R}^{N\times D}$, $D$-dimensional representations for $N$ patches, 
the router network $R(\cdot)$ produces pre-softmax logits
$\mathbf{z}=R(\mathbf{h})\in\mathbb{R}^{N\times E}$, and the gating probabilities are obtained as
\begin{equation}
\mathbf{p}=\mathrm{softmax}(\mathbf{z}).
\end{equation}
For each token representation $\mathbf{h}_b \in \mathbb{R}^d$, 
the router produces gating probabilities 
$\mathbf{p}_b = \mathrm{softmax}(R(\mathbf{h}_b)) \in \mathbb{R}^E$. 
The top-$K$ experts with the highest probabilities are selected, 
and their outputs are aggregated as a weighted sum:
\begin{equation}
\mathrm{MoE}(\mathbf{h}_b) = 
\sum_{e \in \textsc{TopK}(\mathbf{p}_b)} 
\mathbf{p}_{b,e}\, f_e(\mathbf{h}_b),
\end{equation}
where $\textsc{TopK}(\mathbf{p}_b)$ denotes the indices of the $K$ experts 
with the highest gating probabilities for token $b$.

\begin{algorithm}[t]
\caption{Training procedure of TGR-MoE}
\label{alg:tgr}
\SetAlgoLined
\KwIn{
Pretrained dense teacher model $\mathcal{M}_{\mathrm{teacher}}$, \\
\hspace{1.5em}MoE student model $\mathcal{M}_{\mathrm{student}}$, dataset $\mathcal{D}$, \\
\hspace{1.5em}set of MoE layers $\mathcal{S}_{\mathrm{MoE}}$
}
\KwOut{ Trained student model $\mathcal{M}_{\mathrm{student}}$ }
\For{each mini-batch $(x, y) \in \mathcal{D}$}{
    Forward $\mathcal{M}_{\mathrm{student}}$ and obtain routing probabilities $\{\mathbf{p}^{(i)}\}_{i \in \mathcal{S}_{\mathrm{MoE}}}$\;
    Extract teacher intermediate features $\{\mathbf{h}_t^{(i)}\}$\;
    Compute teacher routing probabilities:
    \[
        \mathbf{p}_t^{(i)}=\mathrm{softmax}\!\left(R_t^{(i)}(\mathbf{h}_t^{(i)})\right).
    \]
    Compute teacher router loss w.r.t. Eqn. \ref{eqn:teacherloss};
    Compute router distillation loss w.r.t. Eqn. \ref{eqn:distilloss};
    Compute student final loss w.r.t. Eqn. \ref{eqn:studentloss};
    Update $\mathcal{M}_{\mathrm{student}}$ and $R_t$ using
    $\mathcal{L}_{\mathrm{student}}$ and $\mathcal{L}_{\mathrm{teacher}}$\;
}
\end{algorithm}

\paragraph{Load-Balancing Loss.}
A major challenge in MoE training is expert imbalance, where one or a few experts dominate computation while others remain inactive.
Following VMoE~\cite{riquelme2021scaling}, an auxiliary load-balancing loss is introduced to encourage even expert utilization.
Given a patch of $N$ tokens with routing probabilities $\mathbf{p}\in\mathbb{R}^{N\times E}$, 
the per-expert importance is defined as
\begin{equation}
\text{Imp}_e(\mathbf{p}) = \frac{1}{N} \sum_{i=1}^{N} \mathbf{p}_{i,e}.
\end{equation}
The load loss $\mathcal{L}_{\text{load}}(\mathbf{p})$ minimizes the coefficient of variation of the expert importances:
\begin{equation}
\mathcal{L}_{\text{load}}(\mathbf{p})
= \left(\frac{\mathrm{std}(\text{Imp}(\mathbf{p}))}{\mathrm{mean}(\text{Imp}(\mathbf{p}))}\right)^2 \propto \mathrm{var}(\text{Imp}(\mathbf{p})),
\end{equation}
where $\text{Imp} = \{ \text{Imp}_e \}_{e=1}^{E}$.
In VMoE, two balancing terms are introduced—the \emph{load loss} and the \emph{importance loss}—which together promote uniform expert utilization. 
For simplicity, we refer to their combination as the overall load loss $\mathcal{L}_{\text{load}}(\mathbf{p})$ throughout this paper.

The overall objective for the VMoE model is expressed as
\begin{equation}
\mathcal{L}_{\text{total}} =
\mathcal{L}_{\text{task}} +
\sum_{i \in \mathcal{S}_{\text{MoE}}}
\lambda_{\text{load}} \, \mathcal{L}_{\text{load}}(\mathbf{p}^{(i)}),
\end{equation}
where $\lambda_{\text{load}}$ is a balancing coefficient for the load loss,
$\mathcal{S}_{\text{MoE}}$ denotes the set of layers containing MoE modules,
and $\mathbf{p}^{(i)}$ represents the routing distribution produced by the router at the $i$-th MoE layer.


\subsection{Teacher-Guided Routing for MoE (TGR-MoE)}
\paragraph{Teacher Router Construction}
To leverage a non-MoE dense teacher model to guide the student's routers, we add auxiliary {\it teacher routers}.
The teacher routers are dedicated to providing pseudo-supervision for student routers, and do not perform actual routing in the sense of MoE.

We first extract intermediate representations $\mathbf{h}_t$ from a pretrained dense teacher model. 
Using these features, we define a teacher router $R_t(\cdot)$ that outputs the expert assignment probabilities:
\begin{equation}
\mathbf{p}_t = \mathrm{softmax}(R_t(\mathbf{h}_t)) \in \mathbb{R}^{N \times E}.
\end{equation}

The teacher router is optimized to produce stable and well-distributed routing using two loss terms: 
the load-balancing loss $\mathcal{L}_{\text{load}}(\mathbf{p})$ and the entropy loss $\mathcal{L}_{\text{ent}}(\mathbf{p})$.
The load-balancing loss encourages uniform utilization of experts and prevents underused experts, 
while the entropy loss prevents the router from collapsing into an overly uniform distribution, 
thereby encouraging confident and distinct expert assignments. 
The entropy loss is defined as:
\begin{equation}
\mathcal{L}_{\text{ent}}(\mathbf{p}) =
- \frac{1}{N} \sum_{i=1}^{N} \sum_{e=1}^{E} \mathbf{p}_{i,e} \log \mathbf{p}_{i,e}.
\end{equation}

The pretrained dense teacher model is frozen, and only the teacher router is optimized.
We optimize the teacher router using a subset of training samples $\mathcal{S}_{\text{MoE}}$ to obtain a stable and balanced routing distribution.
The objective function is defined as:
\begin{equation}\label{eqn:teacherloss}
\mathcal{L}_{\text{teacher}} =
\sum_{i \in \mathcal{S}_{\text{MoE}}}
\left(
\lambda_{\text{load}} \mathcal{L}_{\text{load}}(\mathbf{p}_t^{(i)}) +
\lambda_{\text{ent}} \mathcal{L}_{\text{ent}}(\mathbf{p}_t^{(i)})
\right),
\end{equation}
where $\lambda_{\text{load}}$ and $\lambda_{\text{ent}}$ are balancing coefficients.
This optimization does not aim to minimize the downstream task loss directly. Instead, it constructs a teacher router that promotes desirable routing behaviors in Sparse MoE models, such as balanced expert utilization and the avoidance of overly uniform assignments.
It leverages the pretrained teacher's semantically structured feature space, whose rich and organized representations provide a strong prior for guiding the router toward consistent and meaningful expert assignment.

\paragraph{Teacher-Guided Router Training}

The student router $\mathcal{R}_{\text{student}}$ in the MoE model learns to imitate the teacher router's routing distribution. 
Let $\mathbf{p}_{\text{t}}$ and $\mathbf{p}$ denote the output distributions of the teacher and student routers, respectively. 
We define the distillation loss as:
\begin{equation}\label{eqn:distilloss}
\mathcal{L}_{\text{distill}}(\mathbf{p}, \mathbf{p}_t) =
\mathrm{KL}\!\left(
  \text{stopgrad}(\mathbf{p}_t) \; \| \; \mathbf{p}
\right),
\end{equation}
where $\mathrm{KL}(\cdot\|\cdot)$ denotes the Kullback–Leibler divergence \cite{kullback1951information} and 
$\text{stopgrad}(\cdot)$ indicates the stop-gradient operation, 
which detaches the teacher's output from the computation graph to prevent gradient backpropagation into the teacher network.

The overall objective for the MoE model is then expressed as:
\begin{equation}\label{eqn:studentloss}
\mathcal{L}_{\text{student}} =
\mathcal{L}_{\text{task}} +
\frac{\lambda_{\text{distill}}}{|\mathcal{S}_{\text{MoE}}|}
\sum_{i \in \mathcal{S}_{\text{MoE}}}
\mathcal{L}_{\text{distill}}(\mathbf{p}^{(i)}, \mathbf{p}_t^{(i)})
\end{equation}
where $\mathcal{L}_{\text{task}}$ is the main task loss (\textit{e.g.}, classification), 
and $\lambda_{\text{distill}}$ is a hyperparameter controlling the distillation strength.

In practice, we jointly train the teacher router together with the student MoE model.
Since the teacher's backbone is frozen and only the router receives its own auxiliary losses,
the teacher routing converges quickly and remains stable throughout training.
We also observed that pretraining and freezing the teacher router in advance yields nearly identical final accuracy.
Therefore, for simplicity and efficiency, we adopt the joint training scheme in all experiments.
The overall training algorithm for TGR-MoE is summarized in Figure~\ref{fig:overview} and Algorithm~\ref{alg:tgr}.

%% file: sec/4_evaluation.tex
\section{Results}
\label{sec:evaluation}

\subsection{Implementation Details}
We adopted the DeiT \cite{pmlr-v139-touvron21a} architecture as the backbone to ensure a fair comparison, since our proposed TGR-MoE employs a teacher-model-based training paradigm and DeiT offers a well-established and efficient framework for incorporating teacher signals in vision transformers.
To maintain fairness, all experiments, including both the proposed TGR-MoE and the baseline VMoE \cite{riquelme2021scaling} employed the same DeiT-style distillation strategy applied to the final classification layer.  
The MoE configuration followed the original VMoE setup.
The detailed settings are provided in Table~\ref{tab:params}.
As the teacher model, we selected DeiT-III \cite{deit3} pretrained on ImageNet-21K~\cite{imagenet21k} because it achieves strong top-1 accuracy while sharing the same architectural family as the student models. This architectural alignment reduces the inductive gap between teacher and student, facilitating more stable and effective knowledge transfer.

Following the DeiT architectures, which consist of 12 transformer layers, 
the 8th, 10th, and 12th layers were replaced with MoE layers 
across all model variants (Tiny, Small, and Base).
Since the DeiT-III teacher model does not include a distillation token,
we use the routing output associated with the teacher's CLS token
as the distillation target for the student router.
All models were trained on ImageNet-1K~\cite{imagenet} with an input resolution of $224\times224$ 
and fine-tuned on CIFAR-10, CIFAR-100~\cite{cifar}, and Oxford-IIIT Pets~\cite{pets} in later experiments.  
We adopted AdamW optimizer with cosine learning rate scheduling, 
and data augmentations (RandAugment~\cite{cubuk2019randaugment}, Mixup~\cite{zhang2018mixup}, CutMix~\cite{yun2019cutmix}) applied consistently across all methods. Experiments were conducted on 2× or 4× H100 GPUs depending on the model size. Note that we trained all student models from scratch. Some experiments were run only on the Tiny and Small variants due to computational limits.
Detailed hyperparameter settings are provided in Supplementary Section A.

\begin{table}[t]
\centering
\caption{Model configurations and loss coefficients used in all experiments. $E$ denotes the number of experts.}
\scalebox{0.90}{
\begin{tabular}{lcccc}
\toprule
\textbf{Model} & \textbf{$E$} & \textbf{Teacher} & \textbf{Accuracy (Teacher)} \\
\midrule
Tiny  & 16 & DeiT-III-Small & 83.1\% \\
Small, Base & 8  & DeiT-III-Base  & 85.7\% \\
\midrule
\multicolumn{5}{l}{
\textit{Loss coefficients:} 
$\mathcal{L}_\text{load}=0.005$, 
$\mathcal{L}_\text{distill}=5.0$, 
$\mathcal{L}_\text{ent}=0.005$
} \\
\bottomrule
\end{tabular}
}
\label{tab:params}
\end{table}

\begin{table*}[t]
\centering
\caption{
Comparison of Top-1 accuracy (\%) across datasets and model scales.
For V-MoE and the proposed TGR-MoE, results are reported as K=1 / K=2 (left / right) in top-K routing.
Models with K = 1 and K = 2 are trained independently.
For all other baselines without a slash, the reported values correspond to K=1 or an equivalent computational cost.
}
\vspace{-1mm}
\begin{tabular}{clccccc}
\toprule
\multirow{2.3}{*}{Model Size} & \multicolumn{1}{c}{\multirow{2.3}{*}{Model}} & \multicolumn{5}{c}{Top-1 Accuracy (\%)} \\
\cmidrule(lr){3-6}
& & ImageNet-1K~\cite{imagenet} & CIFAR-10 & CIFAR-100~\cite{cifar} & Pets~\cite{pets} \\
\midrule
\multirow{6}{*}{Tiny} 
 & ViT (dense) & 74.62 & 97.78 & 85.43 & 89.86 \\
 & VMoE~\cite{riquelme2021scaling} & 77.85 / 78.21 & 97.98 / 97.72 & 86.20 / 85.75  & 89.82 / 90.17 \\
 & VMoE w/ z-loss~\cite{zoph2022stmoe} & 77.99   & 97.84  &  86.13  & 90.15  \\
 & Expert Choice MoE~\cite{NEURIPS2022_2f00ecd7} & 77.82   & 97.83   & 86.60   & 91.21  \\
 & SoftMoE~\cite{puigcerver2023sparse}  & \textbf{79.31}  & 97.99  & 86.80  & \textbf{91.91}  \\
 & \textbf{TGR-MoE (ours)} & 78.78 / 79.24 & \textbf{98.27 / 98.37} & \textbf{87.03 / 86.87} & 91.78 / 91.63 \\
\addlinespace[0.4em]
\midrule
\multirow{6}{*}{Small} 
 & ViT (dense) & 81.74 & 98.35 & 88.60 & 92.75 \\
 & VMoE  & 82.63 / 82.81 & 98.51 / 98.68 & 88.68 / 88.98 & 93.31 / 93.10 \\
 & VMoE w/ z-loss &  82.42 & 98.62  & 89.36  & 93.04 \\
 & Expert Choice MoE &  82.09 & 98.55 & 89.21  & 93.69 \\
 & SoftMoE &  82.76  & 98.58 &  88.45 & 93.32 \\
 & \textbf{TGR-MoE (ours)} & \textbf{83.34 / 83.68} & \textbf{98.90 / 98.93} & \textbf{90.26 / 90.48} & \textbf{93.79 / 94.14} \\
\addlinespace[0.4em]
\midrule
\multirow{4}{*}{Base} 
 & ViT (dense) & 84.02 & 98.76 & 89.72 & 93.65 \\
 & VMoE  & 83.97 / 84.08 & 98.66 / 98.78 & 89.04 / 89.26 & 93.79 / 93.45 \\
 & VMoE w/ z-loss  & 84.03  & 98.81  & 89.99  & 94.21  \\
 & \textbf{TGR-MoE (ours) } & \textbf{85.46 / 85.34} & \textbf{98.99 / 99.05} & \textbf{91.07 / 91.07} & \textbf{94.63 / 94.97} \\
\bottomrule
\end{tabular}
\vspace{-2mm}
\label{tab:main_results}
\end{table*}




\subsection{Quantitative Evaluation}
\label{sec:quantitative}
We first conducted pre-training on ImageNet-1K to compare the performance of ViT, VMoE, and our proposed TGR-MoE. In addition to standard VMoE, we evaluated Expert Choice MoE~\cite{NEURIPS2022_2f00ecd7}, z-loss regularization~\cite{zoph2022stmoe}, and SoftMoE~\cite{puigcerver2023sparse}. To ensure a fair comparison, Expert Choice MoE and SoftMoE were configured so that each expert processed, on average, the same number of tokens.

As shown in Table~\ref{tab:main_results}, TGR-MoE achieves consistently strong performance across all model scales. It reliably surpasses ViT and VMoE, and is competitive with or better than Expert Choice MoE and SoftMoE on most datasets. For example, on ImageNet-1K, TGR-MoE improves the Tiny model's Top-1 accuracy from 77.85\% to 78.78\% and the Small model from 82.63\% to 83.34\% under the $K{=}1$ setting. These gains demonstrate that providing stable global guidance during routing yields benefits beyond those achieved by existing capacity-balancing or continuous-routing approaches.



We further applied the same training scheme during fine-tuning, where TGR-MoE continued to show superior performance on downstream datasets such as CIFAR-100.  
These results demonstrate that teacher-guided routing leads to improved performance and more effective expert utilization in both pre-training and fine-tuning.
Additional analysis and comparisons are provided in Supplementary Section C.

\paragraph{Effect of TGR-MoE during fine-tuning.}
To further examine whether the routing structure learned during pre-training is preserved during transfer, we conducted fine-tuning experiments on CIFAR-10, CIFAR-100, and Oxford-IIIT Pets, using Tiny, Small, and Base models pretrained on ImageNet-1K.
We compared two configurations:  
(i) fine-tuning a pre-trained TGR-MoE model under the standard VMoE objective, and  
(ii) continuing to apply TGR-MoE during fine-tuning.

As shown in Table~\ref{tab:ft}, maintaining TGR-MoE during fine-tuning consistently achieves the highest accuracy  
(86.95\%, 90.26\%, and 91.07\% for Tiny, Small, and Base, respectively),  
whereas using the pre-trained TGR-MoE only for initialization yields limited gains  
(86.15\%, 89.18\%, and 90.05\%) compared to the VMoE baselines (86.14\%, 88.68\%, and 89.04\%).  
Despite strong pre-training, the performance converges close to the baseline when TGR-MoE is not applied during fine-tuning,
suggesting that the routing knowledge acquired during pre-training is difficult to retain without explicit guidance.
Applying TGR-MoE throughout fine-tuning helps preserve and refine this routing knowledge,  
leading to more stable expert utilization and consistent transfer performance.

\begin{table}[t]
\caption{
Effect of applying TGR-MoE during fine-tuning ($K{=}1$, CIFAR-100).  
Maintaining TGR-MoE during fine-tuning yields the highest accuracy across all model scales,  
showing that explicit routing guidance preserves pre-trained routing knowledge and stabilizes transfer learning.
}
\vspace{-1mm}
\centering
\scalebox{0.96}{
\begin{tabular}{lccc}
\toprule
\multicolumn{1}{c}{\multirow{2.3}{*}{Setting}} & \multicolumn{3}{c}{Top-1 Accuracy (\%)} \\
\cmidrule(lr){2-4}
 & Tiny & Small & Base \\
\midrule
ViT & 85.43 & 88.60 & 89.72 \\
VMoE & 86.14 & 88.68 & 89.04 \\
TGR-MoE (pretrained only) & 86.15 & 89.18 & 90.05 \\
TGR-MoE (during fine-tuning) & \textbf{86.95} & \textbf{90.26} & \textbf{91.07} \\
\bottomrule
\end{tabular}
}
\label{tab:ft}
\end{table}

\subsection{Effect of Varying Number of Experts} \label{sec:num_experts}

\begin{table*}[t]
\caption{Comparison of Top-1 accuracy with different numbers of Experts (Tiny model, ImageNet-1K).  
The advantage of TGR-MoE becomes more evident as the number of experts increases.}
\vspace{-1mm}
\centering
\begin{tabular}{lccccccc}
\toprule
Number of experts & 1 (Dense) & 4 & 8 & 16 & 32 & 64 & 128 \\
\midrule
VMoE & 74.62\% & 76.18\% & 77.39\% & 77.85\% & 78.41\% & 78.74\% & 79.38\% \\
\textbf{TGR-MoE (ours)} & -- & \textbf{76.59\% } & \textbf{77.81\% } & \textbf{78.78\% } & \textbf{79.35\% } & \textbf{79.95\% } & \textbf{80.36\% } \\
 &  & \textbf{(+0.41\%)} & \textbf{(+0.42\%)} & \textbf{(+1.07\%)} & \textbf{(+0.94\%)} & \textbf{(+1.21\%)} & \textbf{(+0.98\%)} \\
\bottomrule
\end{tabular}
\vspace{-2mm}
\label{tab:expert_scaling}
\end{table*}
We further investigated how the number of experts $E$ influences model performance and training stability.
Increasing $E$ naturally raises the sparsity of the model, as only a small subset of experts is activated for each token.
However, this heightened sparsity also amplifies the difficulty of the routing task, often resulting in unstable or suboptimal expert utilization in conventional MoE training. To assess whether the proposed method can sustain stable learning under such sparse regimes,
we evaluated Tiny-scale models with $E \in \{4, 8, 16, 32, 64, 128, 256\}$.
The results are summarized in Table~\ref{tab:expert_scaling}.

As $E$ increases, both VMoE and TGR-MoE exhibit consistent performance improvements, confirming that expanding expert capacity enhances representational power.
However, the gains realized by VMoE taper off at larger expert counts, whereas TGR-MoE maintains steady improvements across all configurations---achieving a +1.07\% boost at $E{=}16$ and sustaining meaningful gains even at $E{=}128$.
These findings suggest that teacher-guided routing enables more reliable expert allocation as model capacity scales, thereby improving the overall scalability of sparse expert architectures.





%% file: sec/5_analysis.tex
\section{Analytical Discussion}
\label{sec:analysis}
\begin{figure}[t]
  \centering
  \begin{minipage}[t]{1\linewidth}
    \centering
    \includegraphics[width=\linewidth]{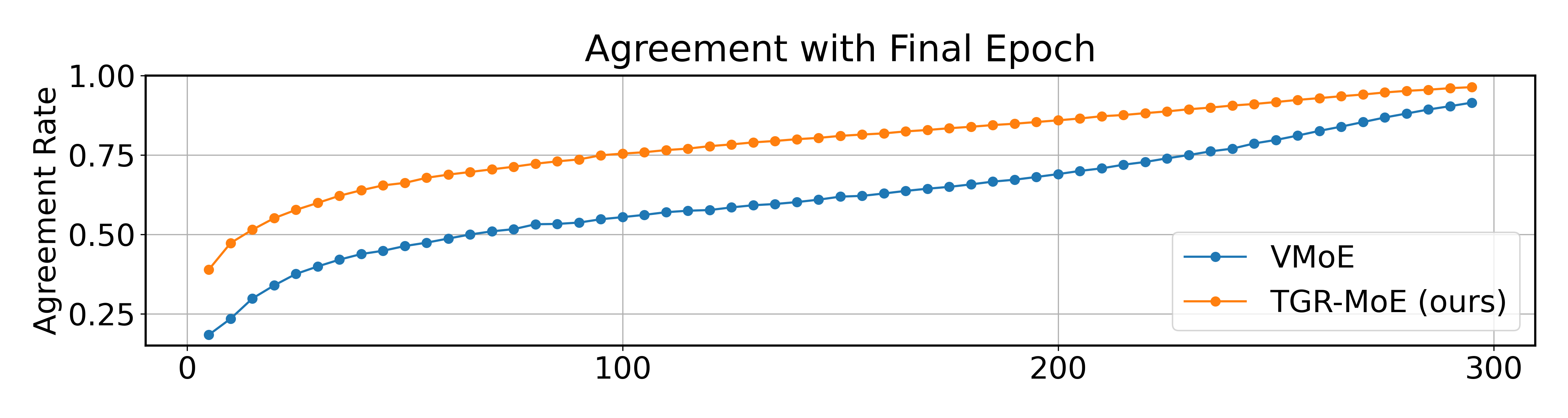}
    \label{fig:consistency_top}
  \end{minipage}

  \vspace{-6.3mm}

  \begin{minipage}[t]{1\linewidth}
    \centering
    \includegraphics[width=\linewidth]{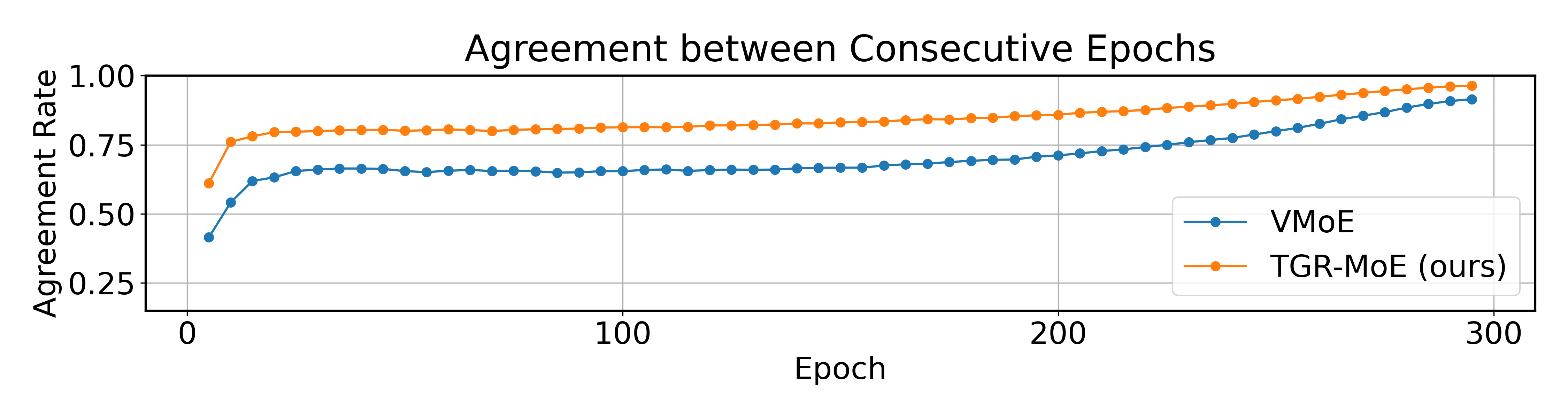}
    \label{fig:consistency_bottom}
  \end{minipage}
  \vspace{-8mm}
  \caption{
    Comparison of routing consistency during training.
    (\textbf{Top}) Agreement rate with the final epoch.
    (\textbf{Bottom}) Agreement rate between consecutive epochs, evaluated every 5 epochs.
    The proposed method significantly reduces routing fluctuations and achieves more stable expert selection, averaged over all MoE layers.
  }
  \label{fig:consistency}
  \vspace{-12mm}
\end{figure}


\subsection{Analysis of Routing Consistency}
\label{sec:analysis_consistency}
To assess how routing behavior evolves during training and whether a method suppresses 
frequent changes in expert selection, we evaluate \emph{routing agreement} of TGR-MoE and VMoE trained on ImageNet-1K in Table~\ref{tab:main_results}, which measures 
the proportion of tokens whose assigned expert matches a reference assignment.

As shown in Figure~\ref{fig:consistency} (top), the proposed method maintains a consistently 
higher agreement with the final model throughout training. Around the midpoint of optimization, 
VMoE still differs from the final routing assignments for roughly 40\% of inputs, indicating 
that its expert selection continues to shift even in later stages. In contrast, our method 
surpasses 70\% agreement within the first 50 epochs and remains comparatively stable thereafter, 
demonstrating that teacher-guided routing enables the router to settle into a consistent usage 
pattern much earlier. A similar trend is observed when examining agreement between consecutive epochs 
(Figure~\ref{fig:consistency}, bottom). Our method maintains a high and steady level of agreement, 
approximately 0.8 throughout training. VMoE, however, exhibits pronounced oscillations, especially 
in the early training phase, with agreement frequently dropping to 0.5--0.6. Overall, these 
results indicate that teacher-guided routing effectively suppresses routing fluctuation.

To further examine whether the routing structure learned during pre-training is preserved after 
transfer, we measured the agreement rate of router outputs before and after fine-tuning on 
CIFAR-100. For the conventional VMoE, agreement dropped to only 50.56\%, indicating that 
fine-tuning substantially altered the routing configuration. In contrast, the proposed TGR-MoE 
maintained a much higher agreement of 73.75\%, suggesting that the teacher-guided routing 
mechanism effectively preserves the pre-trained routing behavior and provides a more consistent 
initialization for downstream adaptation.

\subsection{Interaction of Task and Distillation Losses}
While TGR-MoE trains the student router using both task and distillation losses, the interaction between these two signals remains unclear. We hypothesize that task gradients may conflict with the teacher-guided routing signal, especially in the early stages when routing decisions are still unstable. Meanwhile, it was reported that applying distillation throughout training can impede late-stage convergence, and that restricting it to earlier epochs often improves final performance~\cite{pmlr-v162-rajbhandari22a}. Motivated by this, we investigate how the balance between task-driven and distillation-driven supervision shapes router learning in TGR-MoE. To study this, we compare three training configurations: (1) distillation-only router training, (2) distillation applied only during the first half of training, and  (3) the standard TGR-MoE setup with both losses active throughout.

The results are illustrated in~Table~\ref{tab:taskloss}. Two observations emerge. First, and surprisingly, explicit task supervision is not essential for router optimization: distillation alone achieves competitive accuracy (77.83\%), indicating that the teacher's routing distribution provides sufficiently strong guidance.  
Second, applying distillation only in the early phase yields the best performance (78.13\%), suggesting that teacher-guided signals are most beneficial when routing is volatile, while task gradients become more effective once expert selection has stabilized. These findings reveal that the role of distillation in router training is inherently \emph{phase-dependent}: strong early guidance stabilizes routing, whereas relaxing distillation 
pressure later improves task convergence.

\begin{table}[t]
\caption{
Comparison of different \textbf{TGR-MoE training variants} using the Tiny model with 8 experts on ImageNet-1K.  
While distillation alone provides sufficient supervision for stable routing, 
applying it primarily in the early stage and switching to task optimization later 
yields the best final performance.
}
\centering
\begin{tabular}{lc}
\toprule
Training Configuration & Accuracy (\%) \\
\midrule
Distillation only & 77.83 \\
Distillation (first half) + Task & \textbf{78.13} \\
Distillation + Task & 77.81 \\
\bottomrule
\end{tabular}
\label{tab:taskloss}
\end{table}

\subsection{Analysis of Upper-Bound Routing}
\label{sec:ablation}
To understand the potential benefits of routing based directly on teacher features, we analyze an upper-bound configuration in which the router itself is trained on top of a pretrained dense teacher backbone and used as the routing mechanism at inference time. Although this setup is not feasible in practice as the teacher model cannot be deployed during inference, it serves as an oracle that reveals the maximum achievable benefit when routing 
decisions are made from the teacher's rich feature representations rather than those of the student. Concretely, we train the teacher router using \emph{task} and load-balancing losses. Unlike our proposed TGR-MoE, this configuration does 
not distill teacher routing into a student router; instead, the teacher router directly performs all routing decisions during both training and  inference. This allows us to evaluate the upper bound of expert allocation quality that a student router could aspire to imitate.

Table~\ref{tab:ablation} reports the results for the Tiny model with 8 experts. Using the teacher router for inference yields the highest accuracy (80.19\%), suggesting that routing guided by the teacher's structured and semantically rich feature space can substantially improve expert utilization. This highlights that current student-side MoE routers still leave significant room for improvement in how effectively they leverage expert capacity. 
In contrast, the student router trained solely through imitation of the teacher distribution shows a substantial performance drop (74.84\%). This indicates that it cannot fully reproduce the complex routing behaviors available to the teacher, likely due to limited representational capacity, accumulated approximation errors across layers, and the fact that it never performs routing during training and therefore cannot adjust its behavior based on task-driven signals. Overall, while direct teacher-guided routing represents an unattainable but informative upper bound, TGR-MoE provides a practical middle ground: it transfers meaningful routing structure from the teacher while remaining deployable at inference time, achieving stable and effective routing without requiring access to the teacher model. We provide additional details of this experiment in Supplementary Section B.

\begin{table}[t]
\caption{
Evaluation of the effectiveness of incorporating teacher knowledge into routing.  
Teacher-routed inference performs routing directly with the pretrained teacher during evaluation,
achieving the highest accuracy and illustrating the potential benefit of teacher-informed routing.  
Student-routed inference (w/ distillation) learns to imitate the teacher's routing but performs inference without the teacher model, 
showing a performance drop due to limited adaptation capacity.  
TGR-MoE achieves a balance between these settings, effectively transferring teacher knowledge while remaining teacher-free at inference.
}
\centering
\begin{tabular}{lc}
\toprule
Configuration & Accuracy (\%) \\
\midrule
VMoE baseline & 77.39 \\
TGR-MoE (ours) & 77.81 \\
Student-routed inference (w/ distillation) & 74.84 \\
\addlinespace[0.15em]
\hdashline[2pt/2pt]
\addlinespace[0.15em]
Teacher-routed inference (upper bound) & 80.19 \\
\bottomrule
\end{tabular}
\label{tab:ablation}
\end{table}

\subsection{Further Analysis of Teacher-Guided Routing}
\label{sec:routing_analysis}

To better understand how teacher knowledge stabilizes routing, 
we analyze the behavior of teacher-guided routing in detail. 
We first study which layer of the teacher model provides the most effective routing supervision, 
and then evaluate how faithfully the student router imitates the teacher router.

\subsubsection*{Which teacher layer provides the best guidance?}
We evaluate several choices of teacher feature layers for constructing routing guidance, 
using the Tiny model with 8 experts. 
When the routing signal is generated from the teacher's final-layer features, 
accuracy drops substantially to 75.83\%, considerably lower than our proposed method (77.81\%) 
and even below the standard VMoE baseline. 
This suggests that the high-level representations of a pretrained teacher are too abstract and 
task-specialized to serve as effective routing cues for the student.  
In contrast, layer-aligned intermediate features provide a better structural match, 
yielding a more compatible and effective supervision signal for guiding expert allocation.

\subsubsection*{Agreement between teacher and student routers.}
We next measure how closely the student router replicates teacher routing behavior by computing 
the top-1 expert selection agreement between the two routers across different MoE layers.
As reported in Table~\ref{tab:router_acc}, the agreement consistently improves with model scale, 
indicating that larger student models have greater capacity to approximate the teacher's routing boundaries.  
However, perfect alignment is not achieved even in the Base model, reflecting the inherent 
capacity gap between the student router and the teacher's richer representation space.

Across layers, we observe a monotonic decrease in agreement toward deeper stages. 
This is expected: deeper student representations naturally deviate more from the teacher's, 
and forcing strict alignment in these layers can conflict with task optimization.  
Indeed, teacher routing, while beneficial as an upper bound, is not necessarily optimal for the 
student due to representational mismatch.  
Thus, a balance between imitation and adaptation is required: the student should leverage 
teacher-informed structure where beneficial while retaining flexibility in later layers.  
This trade-off is precisely what TGR-MoE embodies, enabling effective routing transfer without 
over-constraining the student model.

\begin{table}[t]
\caption{
Top-1 expert selection agreement between teacher and student routers across different layers.  
The agreement improves with model size but decreases in deeper layers, 
suggesting that the student router approximates but does not perfectly replicate the teacher's routing behavior.
}
\centering
\begin{tabular}{lccc}
\toprule
\multirow{2.3}{*}{Model} & \multicolumn{3}{c}{Agreement (\%)} \\
\cmidrule(lr){2-4}
 & Layer 8 & Layer 10 & Layer 12 \\
\midrule
Tiny  & 74.81 & 70.60 & 53.43 \\
Small & 80.01 & 75.49 & 59.87 \\
Base  & 81.27 & 77.06 & 67.15 \\
\bottomrule
\end{tabular}
\label{tab:router_acc}
\end{table}



%% file: sec/6_conclusion.tex
\section{Conclusion}
\label{sec:conclusion}
In this work, we introduced TGR-MoE, a teacher-guided routing framework in which a lightweight
router attached to a pretrained dense teacher provides stable and informative
routing distributions as supervision for the student MoE router. Through extensive experiments on ImageNet-1K and multiple downstream
benchmarks, we demonstrated that TGR-MoE consistently improves accuracy across
model scales and remains effective as the number of experts increases,
highlighting its scalability advantages over standard VMoE and other MoE
variants. Our analyses further revealed that TGR-MoE substantially stabilizes
routing during training, preserves routing patterns more faithfully during
fine-tuning, and reduces fluctuations that typically hinder expert
specialization. Importantly, the framework requires access to the teacher model
only during training, incurring no additional inference cost. Overall, our findings indicate that effective MoE routing requires supervisory
signals beyond task-driven gradients, and that incorporating teacher-informed
routing priors offers a simple yet powerful mechanism for achieving stable,
scalable, and high-performing expert-based architectures.

\paragraph{Limitations and Future Work.}
Our experiments were primarily focused on image classification tasks; extending TGR-MoE to natural language and multimodal settings 
remains an important direction for future work.  
Since MoE models are highly sensitive to data scale, future studies should also examine the generalization and scalability of TGR-MoE on larger datasets.  
We believe that such extensions could further validate and enhance the potential of TGR-MoE in large-scale, real-world applications.


%% file: sec/X_suppl.tex
\clearpage
\setcounter{page}{1}
\maketitlesupplementary

\renewcommand{\thefigure}{S\arabic{figure}}
\renewcommand{\thetable}{S\arabic{table}}
\renewcommand{\theequation}{S\arabic{equation}}
\renewcommand{\thesection}{\Alph{section}}


\section{Hyperparameter Details}
\label{sec:hyperparams}

Table~\ref{tab:hyperparams} summarizes all hyperparameters used in our experiments.
Unless otherwise specified, we use identical optimizer settings, data augmentations,
training schedules, and regularization across all baselines to ensure fair comparison.
Architectural configurations, such as depth and sizes of MLPs, follow the standard DeiT~\cite{pmlr-v139-touvron21a} family. For all MoE variants, we use the same configuration as VMoE~\cite{riquelme2021scaling}.
For SoftMoE~\cite{puigcerver2023sparse} and Expert-Choice MoE~\cite{NEURIPS2022_2f00ecd7}, we set the expert capacity (slots) so that each expert
processes approximately $K/E$ of the tokens (where $E$ is the number of experts and $K$ is the number of selected experts per token in VMoE), matching the computation profile of VMoE;
the capacity factor is adjusted accordingly.

\section{Details of the Experiment in Section~5.3}

This experiment evaluates an upper-bound configuration in which the routing
decisions are computed directly from the teacher's backbone features.
Concretely, we attach a router to selected layers of the frozen teacher
(backbone parameters are not updated). Because the teacher representations
encode richer semantic knowledge than those of the randomly initialized
student, the resulting routing is expected to be more stable and more
specialized. This setup allows us to (i) assess whether routing learned from
teacher features is indeed beneficial, and (ii) estimate the maximal accuracy
achievable when expert selection is performed in the teacher's representation
space, providing an upper bound for TGR-MoE.

\medskip
\noindent\textbf{Loss formulation.}
The teacher router is trained using the task loss and a load-balancing loss
summed over all MoE layers:
\begin{equation}
\mathcal{L}_{\text{teacher}}
=
\mathcal{L}_{\text{task}}
+
\lambda_{\text{load}} \sum_{i \in \mathcal{S}_\text{MoE}} \mathcal{L}^{(i)}_{\text{load}} ,
\end{equation}
which is identical to the standard VMoE objective, except that the input
features are the frozen teacher representations.

In this upper-bound configuration, the student model does not perform routing
during training. Its experts are trained solely through distillation from the
teacher router. We therefore separate the losses applied to the student model as:
\begin{equation}
\mathcal{L}_{\text{student-router}}
=
\lambda_{\text{distill}}
\sum_{i \in \mathcal{S}_\text{MoE}}
\mathcal{L}^{(i)}_{\text{distill}},
\end{equation}
\begin{equation}
\mathcal{L}_{\text{student}}
=
\mathcal{L}_{\text{task}}
+
\mathcal{L}_{\text{student-router}} ,
\end{equation}
which makes explicit that the student router receives no task gradient; it is
optimized only via distillation. As a result, the student router is used for the
first time at inference, leading to the observed performance gap reported in
Table~5 of the main paper.

\begin{figure*}[t]
\centering
    \begin{subfigure}[t]{0.32\linewidth}
        \centering
        \includegraphics[width=\linewidth]{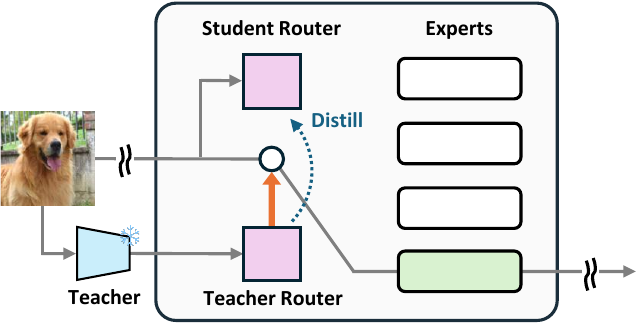}
        \caption{Training with teacher-supervised routing}
        \label{fig:proposed_training}
    \end{subfigure}
    \hfill
    \begin{subfigure}[t]{0.32\linewidth}
        \centering
        \includegraphics[width=\linewidth]{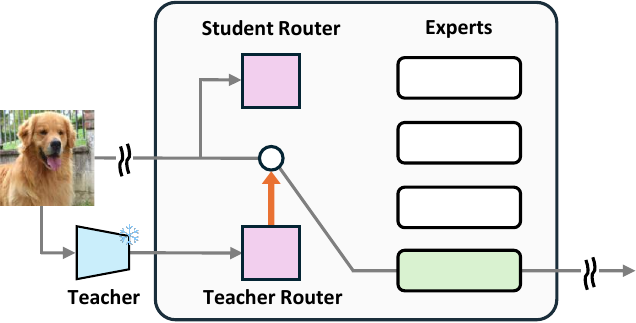}
        \caption{Teacher-routed inference (upper bound)}
        \label{fig:proposed_teacher_infer}
    \end{subfigure}
    \hfill
    \begin{subfigure}[t]{0.32\linewidth}
        \centering
        \includegraphics[width=\linewidth]{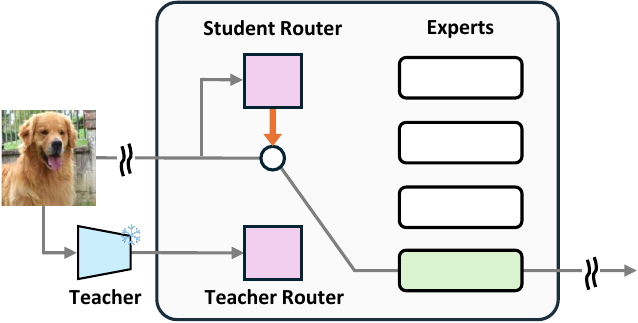}
        \caption{Student-routed inference (with distillation)}
        \label{fig:proposed_student_infer}
    \end{subfigure}
\caption{
Comparison of training and inference configurations used in Section~5.3. 
    (a) During training, routing supervision is provided by the teacher router while the teacher backbone remains frozen. 
    (b) At inference, routing can be performed using the teacher router to obtain an upper-bound performance. 
    (c) Alternatively, routing can be performed by the student router, which was trained only via distillation.
}
\label{fig:upperbound}
\end{figure*}

\medskip
\noindent\textbf{Difference from TGR-MoE.}
The architecture for this experiment is illustrated in Fig.~\ref{fig:upperbound}.
The primary difference from TGR-MoE lies in the \emph{source of routing}:
the teacher router directly selects experts during training and inference,
whereas TGR-MoE uses the teacher router only for supervision, and the student
router performs routing. Furthermore, unlike TGR-MoE, the student router in
this upper-bound configuration receives no task loss and is therefore used
for the first time only at inference. As shown in Table~5 of the main paper,
this mismatch introduces a performance gap, which explains why the student
model cannot fully reach the upper-bound accuracy.

In contrast, TGR-MoE eliminates this gap by always training with the student
router. The router is supervised by the teacher routing prior during
training, but expert selection is ultimately driven by the student router
itself for both training and inference. As a result, TGR-MoE inherits the
teacher's knowledge while maintaining consistency between training
and inference, thereby avoiding the degradation observed in the upper-bound
configuration.

\begin{table*}[t]
\centering
\renewcommand{\arraystretch}{1.25}
\caption{Summary of hyperparameters used in all experiments.}
\vspace{-2mm}
\scalebox{0.93}{
\begin{tabular}{ll}
\hline
\textbf{Item} & \textbf{Setting} \\
\hline
Architecture (DeiT-Tiny) 
& Hidden dim: 192,\ Heads: 3,\ MLP dim: 768,\ Layers: 12,\ MoE layers: \{8,10,12\} \\
Architecture (DeiT-Small) 
& Hidden dim: 384,\ Heads: 6,\ MLP dim: 1536,\ Layers: 12,\ MoE layers: \{8,10,12\} \\
Architecture (DeiT-Base) 
& Hidden dim: 768,\ Heads: 12,\ MLP dim: 3072,\ Layers: 12,\ MoE layers: \{8,10,12\} \\
\hline
Teacher architecture (DeiT-III-Small)~\cite{deit3} 
& Hidden dim: 384,\ Heads: 6,\ MLP dim: 1536,\ Layers: 12,\ Router layers: \{8,10,12\} \\
Teacher architecture (DeiT-III-Base) 
& Hidden dim: 768,\ Heads: 12,\ MLP dim: 3072,\ Layers: 12,\ Router layers: \{8,10,12\} \\
\hline
Image size 
& 224 x 224 \\
Training epochs 
& 300 (ImageNet-1K~\cite{imagenet}),\ 100 (CIFAR~\cite{cifar}, Pets~\cite{pets}) \\
Learning rate 
& $5\times 10^{-4}$,\ warmup from $1\times 10^{-6}$ for first 5 epochs \\
Scheduler 
& Cosine annealing \\
Optimizer 
& AdamW~\cite{adamw} ($\beta_1=0.9$, $\beta_2=0.999$, $\epsilon=1.0\times10^{-8}$,\ weight decay $=0.05$) \\
Data augmentation 
& RandAugment~\cite{cubuk2019randaugment},\ Mixup~\cite{zhang2018mixup},\ CutMix~\cite{yun2019cutmix} \\
\hline
MoE noise 
& Gaussian noise (std $1.0$) added to router logits \\
Load-balancing loss
& $\lambda_{\text{load}}=0.005$ \\
Distillation loss 
& $\lambda_{\text{distill}}=5.0$ \\
Entropy regularization 
& $\lambda_{\text{ent}}=0.005$ (teacher router) \\
\hline
z-loss~\cite{zoph2022stmoe} (comparison) 
& $\lambda_{\text{zloss}}=1.0\times 10^{-3}$ \\
\hline
\end{tabular}
}
\label{tab:hyperparams}
\end{table*}

\begin{figure*}[!t]
  \centering
    \includegraphics[width=\linewidth]{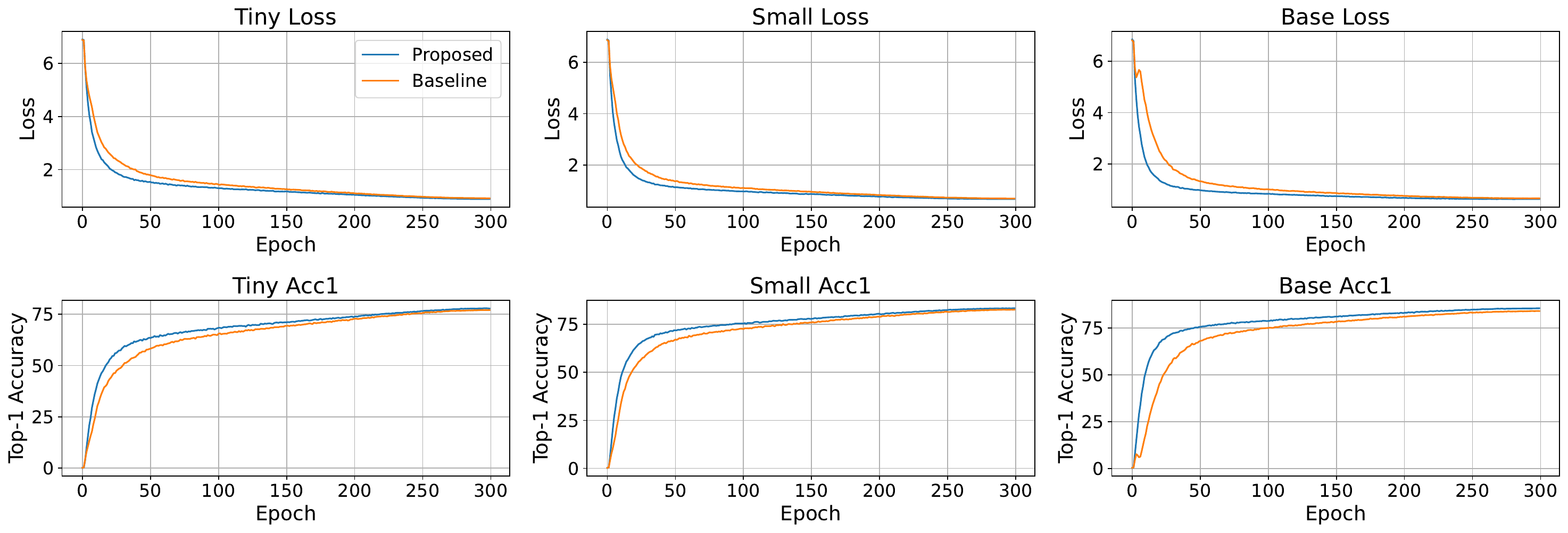}

  \caption{
  Training loss (top row) and training accuracy (bottom row) for the Tiny, Small, and Base models.
  The proposed TGR-MoE consistently converges faster and more stably than VMoE across all model scales.
  }
  \label{fig:supp_training_dynamics}
\end{figure*}


\section{Additional Analysis and Comparison}
\subsection{Training Dynamics of TGR-MoE versus VMoE Across Model Scales}
\label{sec:sup}

To examine how the proposed TGR-MoE stabilizes expert selection and accelerates optimization, we examine the training dynamics of both
TGR-MoE and VMoE during pre-training. In particular, we
compare how quickly the models reduce loss and improve accuracy, as this
reflects the stability of the routing decisions formed in the early stage of
learning.

Figure~\ref{fig:supp_training_dynamics} visualizes the training
accuracy (top row) and training loss (bottom row) for the Tiny, Small, and
Base models (left to right).  
Across all scales, TGR-MoE (blue) exhibits faster convergence than VMoE (orange),
especially at the beginning of training. These results indicate that TGR-MoE
produces more stable training dynamics by mitigating routing instability and
accelerating the formation of well-behaved expert assignments.
\subsection{Analysis of Expert Collapse}
To directly assess whether the proposed method avoids degenerate routing, we report normalized routing entropy in Table~\ref{tab:table}.
The results show that TGR-MoE maintains high expert-utilization entropy across layers and model scales, indicating that the router does not collapse to a trivial or highly imbalanced assignment.
This supports our claim that teacher-guided routing mitigates unstable expert specialization in sparse MoE training.

\subsection{Additional Comparison with Routing Stabilization Baselines}
We additionally compared TGR-MoE with StableMoE~\cite{dai-etal-2022-stablemoe} in the Tiny setting.
StableMoE was originally tested in NLP, but
for verification purposes, we evaluated StableMoE using a Tiny model by copying the routing network, which is then frozen without distillation.
StableMoE achieved \underline{77.19\%} accuracy on ImageNet-1K (8 experts), while ours reached \textbf{77.81\%}
under the same setting.

\subsection{Training Cost and Parameter Overhead}
Table~\ref{tab:table} also reports training time and trainable parameter counts.
The additional cost of TGR-MoE is negligible: the method introduces only lightweight teacher routers and a distillation loss, resulting in almost unchanged training time and parameter count compared with VMoE.
This indicates that the proposed improvement in routing stability does not come with a meaningful computational overhead.

\begin{table}[t]
\centering
\small
\setlength{\tabcolsep}{4.1pt}
\setlength{\abovecaptionskip}{3pt}
\setlength{\belowcaptionskip}{0pt}
\caption{Training time (sec/epoch, single H100), trainable parameters (including teacher routers; excluding the teacher model), and normalized routing entropy.} 
\label{tab:table}
\begin{tabular}{lccccc}
\hline
Model & Time & Params & L8 ent. & L10 ent. & L12 ent. \\
\hline
VMoE-Ti & 1006  & 19.23M & 0.9098 & 0.7776 & 0.9489 \\
TGR-MoE-Ti & 1020  & 19.25M & 0.9701 & 0.9627 & 0.9420 \\
\hline
VMoE-S & 2560 & 47.26M & 0.9924 & 0.9848 & 0.9938 \\
TGR-MoE-S & 2570 & 47.28M & 0.9772 & 0.9708 & 0.9878 \\
\hline
VMoE-B & 4930 & 186.53M & 0.9677 & 0.9906 & 0.9888 \\
TGR-MoE-B & 4954 & 186.55M & 0.9462 & 0.9789 & 0.9892 \\
\hline
\end{tabular}
\end{table}